\documentclass[sigconf]{acmart}

\AtBeginDocument{%
  \providecommand\BibTeX{{%
    \normalfont B\kern-0.5em{\scshape i\kern-0.25em b}\kern-0.8em\TeX}}}

\usepackage{multirow}
\usepackage{color}
\usepackage{graphics}
\usepackage{bbding}

\def\etal{\emph{et al. }}

\renewcommand\footnotetextcopyrightpermission[1]{} 
\begin{document}
\fancyhead{}

\title{ICECAP: Information Concentrated \\ Entity-aware Image Captioning}


\author{Anwen Hu}
\affiliation{%
  \institution{School of Information\\ Renmin University of China}
  }
\email{anwenhu@ruc.edu.cn}

\author{Shizhe Chen}
\affiliation{%
  \institution{School of Information\\ Renmin University of China}
  }
\email{cszhe1@ruc.edu.cn}

\author{Qin Jin}
\authornote{Corresponding Author}
\affiliation{%
  \institution{School of Information\\ Renmin University of China}
  }
\email{qjin@ruc.edu.cn}

\renewcommand{\shortauthors}{Hu et al.}

\begin{abstract}
  Most current image captioning systems focus on describing general image content, and lack background knowledge to deeply understand the image, such as exact named entities or concrete events.
  In this work, we focus on the entity-aware news image captioning task which aims to generate informative captions by leveraging the associated news articles to provide background knowledge about the target image.
  However, due to the length of news articles, previous works only employ news articles at the coarse article or sentence level, which are not fine-grained enough to refine relevant events and choose named entities accurately.
  To overcome these limitations, we propose an Information Concentrated Entity-aware news image CAPtioning (ICECAP) model, which  progressively concentrates on relevant textual information within the corresponding news article from the sentence level to the word level.
  Our model first creates coarse concentration on relevant sentences using a cross-modality retrieval model and then generates captions by further concentrating on relevant words within the sentences.
  Extensive experiments on both BreakingNews and GoodNews datasets demonstrate the effectiveness of our proposed method, which outperforms other state-of-the-arts. The code of ICECAP is publicly available at \url{https://github.com/HAWLYQ/ICECAP}.
\end{abstract}




\maketitle

\section{Introduction}

\begin{figure}[!tbp]
    \centering
    \includegraphics[width=0.9\linewidth]{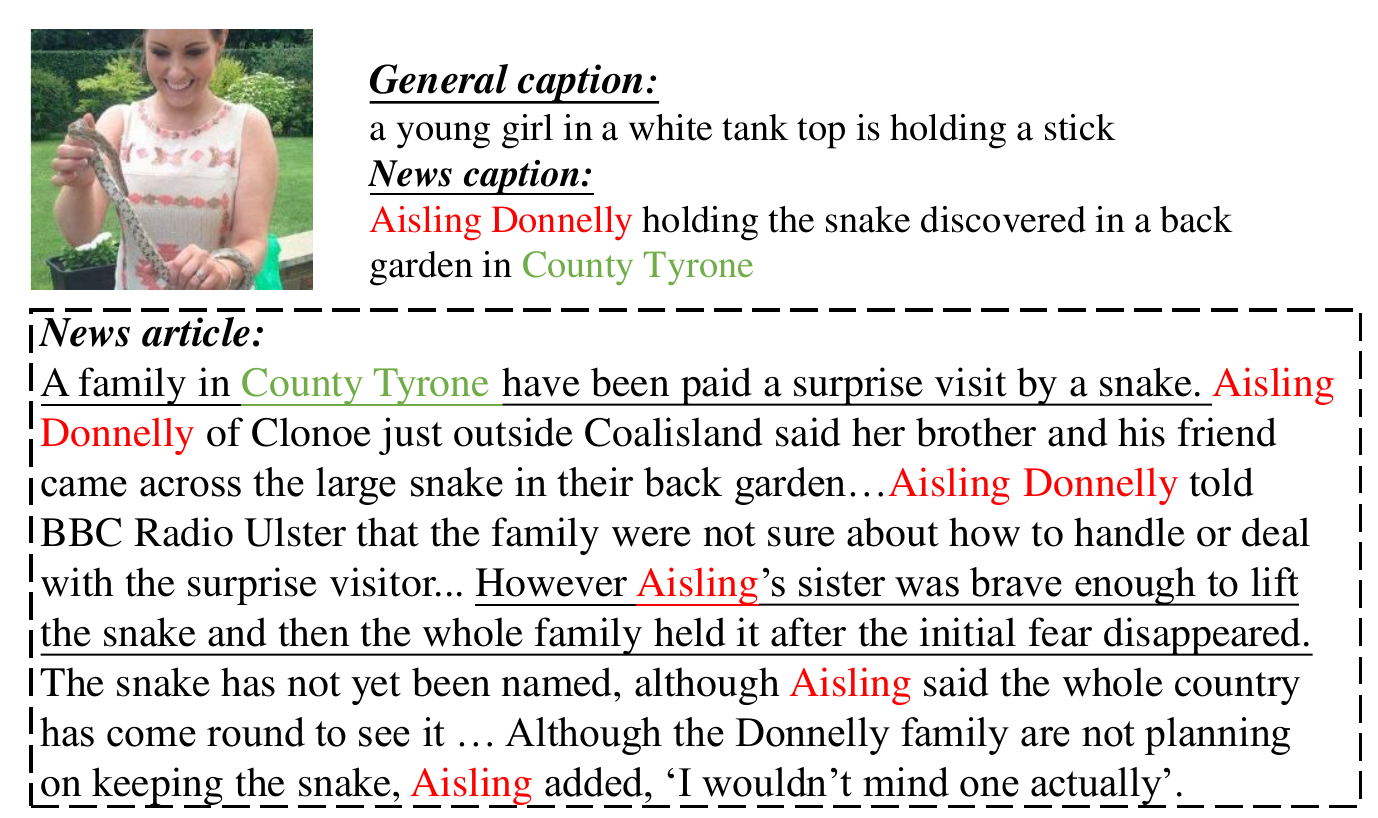}
    \caption{Comparison of the general image caption and entity-aware news image caption. The sentences relevant to the image in the news article are underlined.}
    \label{fig:example}
\end{figure}

How would you describe the image in Figure~\ref{fig:example} to your friends?
If you know nothing about the background of the image, you may simply tell them the general semantic content in the image such as `a young girl in a white tank top is holding a stick'. This is what current image captioning systems \cite{DBLP:conf/icml/XuBKCCSZB15} may do.
However, if you know relevant backgrounds of the image, such as who the girl is, where she is, and why she is doing that, you could leverage such knowledge to generate a more accurate and informative description about the image.
The current machines, nevertheless, lack such abilities to incorporate knowledge in image understanding.
To reduce difficulties of searching for open-domain knowledge, a more constrained task called news image captioning has been proposed, where the prior knowledge is presented in an associated news article along with the image.

There have been a few endeavors to incorporate the textual information from the corresponding news article in order to generate entity-aware image captions. 
For example, \cite{DBLP:journals/tip/TariqF17,feng2012automatic} explored probability models to select relevant texts from the news article. 
Ramisa \etal \cite{ramisa2018breakingnews} propose neural-based model which leverages the article information by introducing a single article feature vector. Later, Biten \etal \cite{DBLP:conf/cvpr/BitenGRK19} propose to attend to different sentences with an attention mechanism when decoding each word.
However, there are two major limitations for generating appropriate named entities in captions in these methods.
Firstly, there can be multiple named entities in an article or even in a single sentence. So leveraging textual information from the news at the article level or even the sentence level is not fine-grained enough for finding appropriate named entities.
Secondly, to overcome the out-of-vocabulary (OOV) problem caused by named entities, previous works \cite{DBLP:conf/emnlp/LuWHJC18,DBLP:conf/cvpr/BitenGRK19} follow a two-stage framework. They first generate the `template caption' which contains the entity types (namely `placeholders'). Then the placeholders are replaced by appropriate entity names during post-processing. Therefore, the selection of the named entities is not optimized during caption generation. Alasdair \etal \cite{Tran2020Tell} propose to represent named entities with byte-pair encoding \cite{DBLP:conf/acl/SennrichHB16a} to resolve the OOV problems, but the byte-pair tokens of a named entity lack actual semantics. 

To overcome these limitations, we propose a novel \textbf{I}nformation \textbf{C}oncentrated \textbf{E}ntity-aware image \textbf{CAP}tioning model (ICECAP)  for News Image Captioning, which progressively concentrates on the relevant textual information in the news article from the sentence level to the word level. 
The proposed approach consists of two main modules: 1) the Information Concentration Module, which realizes the coarse concentration by locating the relevant sentences within the news article based on cross-modal retrieval, and 2) the end-to-end Entity-aware Caption Generation Module, which leverages the textual information of the retrieved sentences at the word level and jointly performs template caption generation and named entity selection. We evaluate the proposed approach on two widely used news image caption datesets. 
The main contributions of this work are as follows:
\begin{itemize}
    \item We propose a novel approach which progressively concentrates on the relevant textual information in the news article for entity-aware image captioning.
    \item The end-to-end caption generation module in our approach can jointly optimize the named entity selection and the template caption generation.
    \item Our proposed method achieves state-of-the-art results on both BreakingNews and GoodNews datasets. 

\end{itemize}

\section{Background}

Given an image $I$ and its associated news article $D=\{S_1, \cdots, S_{N_s}\}$ where $S_i$ is the $i$-th sentence and $N_s$ is the number of sentences, the task of entity-aware news image captioning aims to generate an informative caption $C=\{c_{1}, c_{2}, ..., c_{N_c}\}$ to describe the image $I$, where ${N_c}$ is the number of words.
The caption $C$ should leverage the contextual information in $D$ to understand $I$ at a high interpretation level such as describing more concrete events and named entities in the image.

\noindent\textbf{Article-level Approach.}
In order to integrate document information, Ramisa \etal \cite{ramisa2018breakingnews} encode $D$ into a fixed-dimensional article representation and concatenate it with image features to generate captions in the vanilla encoder-decoder image captioning framework \cite{DBLP:conf/icml/XuBKCCSZB15}.
Although the fused multimodal representation is superior to either of the single modality, the global article representation is hard to capture visual-relevant fine-grained information.
Therefore, it is insufficient to generate concrete events and named entities that may even be out-of-vocabulary words.

\noindent\textbf{Sentence-level Approach.}
To tackle the above limitations, Biten \etal \cite{DBLP:conf/cvpr/BitenGRK19} propose a two-stage pipeline based on attentional image captioning model \cite{DBLP:conf/cvpr/VinyalsTBE15}, including a \emph{template caption generation} stage and a \emph{named entity insertion} stage.

In the \emph{template caption generation} stage, the captioning model integrates a document and an image to generate a template caption $C^p=\{c^{p}_{1}, c^{p}_{2}, ..., c^{p}_{N_p}\}$, where $C^p$ is created by replacing named entities in $C$ with their respective tags.
For example, the template sentence for sentence `Aisling Ddonnelly is discovered in County Tyrone' is `$<$PERSON$>$ is discovered in $<$GPE$>$' (denoting geographic place).
The model encodes document $D$ into a set of sentence-level representations via average pooling word embeddings in each sentence, which are more fine-grained than document encoding in the article-level method.
At decoding phase, the model dynamically attends to both sentence features in the document and region features in the image, and concatenate them as input for the LSTM decoder to predict each word.

Then the \emph{named entity insertion} stage applies post-processing to convert placeholders in template caption into named entities mentioned in the document.
Firstly it ranks sentences in the article, and then selects named entity whose entity category is consistent with the placeholder following the order of appearance in the ranked sentences.
Two strategies are proposed for sentences ranking.
One is Context-based Insertion (\textbf{CtxIns}), which ranks sentences according to cosine distances with the generated template caption.
Another strategy is based on sentence attention mechanism (\textbf{SAttIns}) in the first stage, which ranks sentences by sentence attention scores when generating the placeholder. 

Though the sentence-level approach has improved news image captioning performance, it is hard for the model to learn sentence relevancy from scratch to guide template caption generation, since most sentences in the article can be irrelevant to the input image.
Furthermore, the named entity insertion approaches are ad-hoc without considering fine-grained matching between the placeholder and named entities in the article.
Therefore, the sentence-level approach is still not very  effective in leveraging contextual information in the document to generate news image captions.


\begin{figure*}
    \centering
    \includegraphics[width=0.9\linewidth]{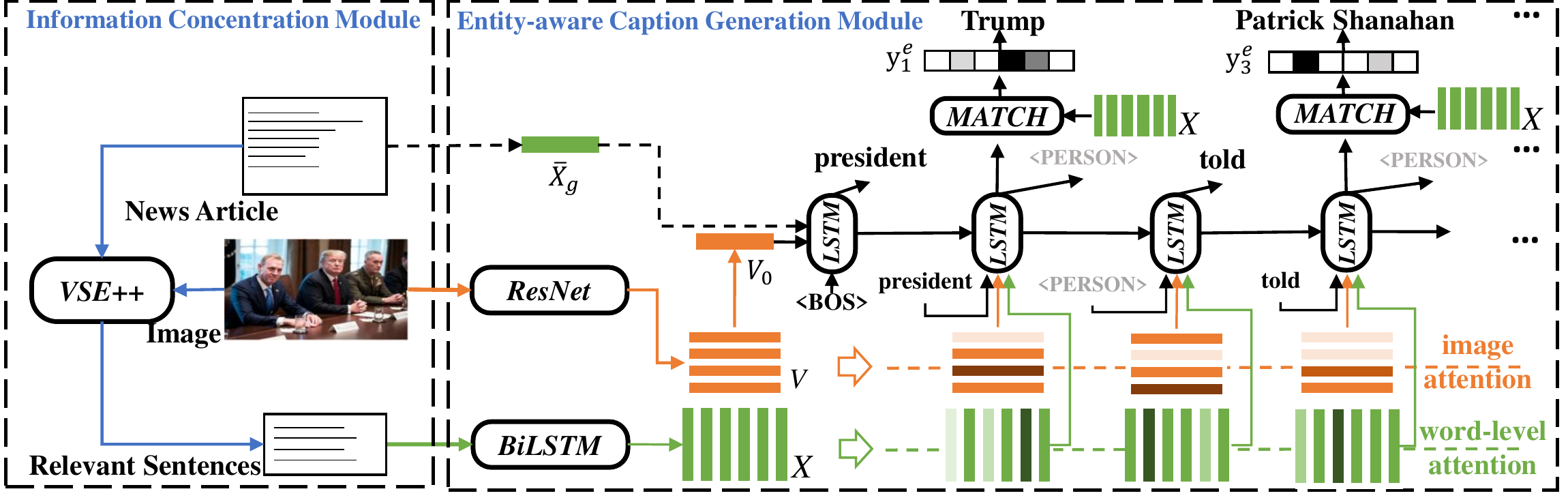}
    \caption{ICECAP framework: Information Concentrated Entity-aware image CAPtioning. Information Concentration Module selects relevant sentences from the article. Entity-aware Caption Generation module generates the entity-aware image caption.}
    \label{fig:overall_arch_match}
\end{figure*}

\section{Information Concentrated Entity-aware Image Captioning}
In this section, we introduce the Information Concentrated Entity-aware image CAPtioning model (ICECAP) for news image caption generation.
As illustrated in Figure~\ref{fig:overall_arch_match}, the ICECAP model consists of two key modules, namely \emph{information concentration} and \emph{entity-aware caption generation}.
The information concentration module firstly selects relevant sentences in the news article for the given image, so that it reduces irrelevant noisy information in the document for later processing and improves efficiency.
Then the entity-aware caption generation module jointly optimizes template caption generation and named entity selection in an end-to-end fashion, which enables to generate named entities from the article more accurately.
We describe the two modules of our ICECAP model in details in following subsections.

\vspace{-8pt}
\subsection{Information Concentration Module}
\label{retrival_sec}
When journalists organize a news article, an attached news image is usually relevant to only a few sentences in the article.
Therefore, concentrating on such relevant sentences from the entire news article can largely reduce the searching scope of useful contextual information for subsequent processing.

In order to select sentences in document $D$ relevant to the given image $I$, the mainstream approach is to use an image-text cross-modal retrieval model, which measures cross-modal similarities in a common embedding space.
In this work, we employ the state-of-the-art VSE++ model \cite{DBLP:conf/bmvc/FaghriFKF18} for image-text cross-modal retrieval. It is impractical to simply apply image-text retrieval models pre-trained on some generic image caption datasets such as MSCOCO \cite{DBLP:conf/eccv/LinMBHPRDZ14} due to large domain discrepancy.
Thus, we need to train image-text retrieval models based on news image captioning dataset BreakingNews \cite{ramisa2018breakingnews} and GoodNews \cite{DBLP:conf/cvpr/BitenGRK19}.
However, there are no available image-sentence training pairs in these two datasets and collecting such manual annotation is costly. Taking into account that the writing style of news image caption is similar to the sentences in the corresponding article, we use the groundtruth caption as the positive sentence for each image and captions describing other images as negative sentences.
Since it is hard and less generalized to directly associate an image with exact named entities, we convert all sentences into template caption format, where all named entities in the sentence are substituted by their entity categories.

The VSE++ model employs CNN to project image $I$ into vector $v$ and LSTM to project sentence $S^{p}$ into vector $s^p$ in a joint embedding space. 
It pushes the distance between positive examples and negative examples larger than certain margin during training, with loss function as follows:
\begin{equation}
    L_r = [\Delta+f_s(v,\bar{s^{p}})-f_s(v, s^{p})]_{+} + [\Delta+f_s(\bar{v}, s^{p})-f_s(v, s^{p})]_{+}
\end{equation}
where $[x]_+ := \mathrm{max}(x,0)$, $\Delta$ is the margin, $(v, s^p)$ is a positive pair while $(v, \bar{s}^p)$ and $(\bar{v}, s^p)$ are two hardest negative pairs for $v$ and $s^p$ respectively in a mini-batch, $f_s(x,y)$ is the cosine similarity.

In the inference phase, we use the VSE++ model to rank sentences in the associated article for each image. We choose top $R$ sentences as relevant ones for subsequent caption generation.

\subsection{Entity-aware Caption Generation Module}
\label{generate_sec}
The main difference of news image captioning compared with generic image captioning is that the news article provides more contextual knowledge for the image such as concrete events and named entities.
Therefore, besides encoding image features into region-level visual representations $V=\{v_{1},v_{2},...,v_{m}\}$ for caption generation, the model also needs to effectively encode articles and leverage article contexts to generate more detailed template caption and insert named entities.

For this purpose, the entity-aware caption generation module firstly encodes articles at global and local levels, where the local encoding benefits from the information concentration module to keep fine-grained word-level information.
Then template caption is generated via dynamically attending to encoded image and word-level article contexts.
Named entities are simultaneously inserted into the template caption via word-level matching, which can be optimized in an end-to-end way to improve accuracy. 

\paragraph{\textbf{Global and Local Article Encoding.}}
Assuming $\{S_1, \cdots, S_{N_s}\}$ are all sentences in the article $D$ and $\{\hat{S}_1, \cdots, \hat{S}_{R}\}$ are sentences relevant to the image $I$ selected by our information concentration module from $D$, we aim to encode the article at both global and local levels for news image captioning.
The goal of global encoding is to utilize full contexts in the article without information loss, while the local encoding enables model to focus on more fine-grained information relevant to the image.

For the global encoding, we employ pre-trained Glove word embedding\cite{DBLP:conf/emnlp/PenningtonSM14} to encode each word in $D$.
Then we obtain sentence embeddings via smooth inverse frequency (SIF) algorithm\cite{DBLP:conf/iclr/AroraLM17}.
The global article feature $\bar{X}_{g}$ is the sum of all sentence embeddings in the article. 

For the local encoding, we only encode relevant sentences after information concentration which can largely reduce computational costs without useful information loss.
We concatenate $\{\hat{S}_1, \cdots, \hat{S}_{R}\}$ into a long word sequence in sequential order to keep temporal transition among these sentences.
The named entities should be replaced because the OOV problem they caused can easily lead to incoherence of sentence semantics.
In this work, we propose an \textbf{entity-distinguished embedding} strategy which replace each named entity with its entity category and a serial number.
For example, if `Joshua' is the first person name appeared in the  article, its all occurrences are replaced by the `PERSON-1' placeholder. Therefore, different named entities of the same entity category can also be distinguished.
The entity-distinguished embedding are randomly initialized while embeddings of other words are initialized as Glove word embeddings.
We utilize BiLSTM model to encode the long word sequence to obtain local word-level features $X=\{x_1, \cdots, x_{N_x}\}$.

\paragraph{\textbf{Template Captioning with Word-level Attention.}}
The decoder is an LSTM model which leverages both article and visual information to generate template captions.

At the beginning of decoding, we fuse global article feature and the average pooling of region-level visual features, to initialize the decoder, which captures all available information at a coarse-grained level.

Then, the decoder dynamically pays attention to different words in the local article encoding and regions in the visual encoding respectively to predict each word.
The word-level attention at step $t$ is calculated as follows:
\begin{gather}
    s_{t}^{i} = \omega^{T}_{X}(\tanh(W_{h}^{X}h_{t-1}+W_{x}^{X}x_{i})),\\
    \alpha_{t} = {\rm Softmax}([s_{t}^{1}, s_{t}^{2}, ..., s_{t}^{N_x}]), \\
    \hat{x}_{t} = \alpha_{t}X \label{eqn_word_att}
\end{gather}
where $x_{i}\in X$, $W_{h}^{X}$, $W_{w}^{X}$, $\omega^{T}_{X}$ are learnable parameters, $\alpha_{t}$ is the word-level attention weights. So $\hat{x}_{t}$ is the attended textual feature at step $t$.
Similarly, we can generate relevant visual feature $\hat{v}_t$ by applying another attention calculation on $h_{t-1}$ and $V$.
We concatenate embedding of the previous generated word $y_{t-1}^{p}$ and $\hat{v}_t, \hat{x}_t$ as the input for LSTM to predict the word, which is:
\begin{gather}
    h_{t} = {\rm LSTM}([W_{y}y_{t-1}^{p}; \hat{x}_{t}; \hat{v}_t], h_{t-1}), \\
    y_{t}^{p} = {\rm Softmax}(W_{h}^{p}h_{t})
\end{gather}
where $W_{y}$ is the word embedding matrix, $W_{h}^{p}$ is learnable parameter, and $[;]$ denotes vector concatenation. 

\paragraph{\textbf{Entity Selection with Word-level Matching.}}
The context of named entities is a reliable indicator to select appropriate named entity from relevant sentences for the placeholders in the template caption.
Therefore, we propose a match layer which compares the hidden state $h_{t}$ with each word-level contextual feature $x_i$ in $X$ to select a named entity. The calculation in the match layer is defined as:
\begin{gather}
    e_{t}^{i} = \omega^{T}_{A}(\tanh(W_{h}^{A}h_{t}-W_{x}^{A}x_{i})),\\
    y_{t}^{e} = {\rm Softmax}([e_{t}^{1}, e_{t}^{2}, ..., e_{t}^{N_x}]), 
\end{gather}
where $W_{h}^{A}$, $W_{x}^{A}$, $\omega^{T}_{A}$ are learnable parameters. $y_{t}^{e}$ denotes the  match probabilities among all words in the relevant sentences for the $t^{th}$ token in the template caption.

\paragraph{\textbf{Training and Inference.}}
During training, we apply the cross-entropy loss for the template caption generation:
\begin{gather}
    {\rm L_{p}} = -\sum_{t=1}^{l}{\hat{y_{t}}^{p} \log(y_{t}^{p})},
\end{gather}
where $\hat{y_{t}}^{p}$ is the true distributions (one-hot vector) for the $t^{th}$ token in the template caption.

For word-level named entity matching, since the target named entity might occur in multiple positions with different contexts in relevant sentences, we do not have the groundtruth of the exact position where its context is the best for the placeholder in the template caption.
Therefore, we treat all occurred positions of the named entity equally as our matching target $\hat{y_{t}}^{e}$ for the $t^{th}$ token. The value of the corresponding position of its occurrence is set as 1 in $\hat{y_{t}}^{e}$. 
For non-entity tokens or entity tokens that fail to occur in relevant sentences, the target $\hat{y_{t}^{e}}$ is a zero vector.
The loss for named entity selection is calculated as:
\begin{gather}
    {\rm L_{e}} = -\sum_{t=1}^{l}{\hat{y_{t}^{e}} \log(y_{t}^{e})},
\end{gather}

We combine the template caption generation loss and the named entity selection loss with a manually defined trade-off scalar $\gamma$:
\begin{gather}
    {\rm L} = {\rm L_{p}} + \gamma {\rm L_{e}}
    \label{final_loss}
\end{gather} 

During inference, we apply greedy search to decode the template caption $C^{p}=(c_{1}^{p}, c_{2}^{p}, ..., c_{l}^{p})$. If $c_{i}^{p}$ is a placeholder, we use the normalized match scores $\boldsymbol{y_{i}^{e}}$ to locate entity name $c_{i}$ with corresponding entity category in $C^p$ from the relevant sentences.

\begin{table}[!htbp]
    \centering
    \caption{Partition Statistics of BreakingNews and GoodNews.}
    \label{tab:dataset statistic}
    \vspace{-10pt}
    \begin{tabular}{p{0.2\linewidth}p{0.15\linewidth}p{0.13\linewidth}p{0.12\linewidth}p{0.12\linewidth}}
    \toprule
    \textbf{Dataset} & \textbf{category}  & \textbf{\#Train}  & \textbf{\#Val} & \textbf{\#Test} \\
    \midrule
    \multirow{2}*{BreakingNews}    & Image     & 65,542    & 21,705        & 22,216 \\
    ~                               & Article   & 56,118    & 18,737        & 19,111 \\
    \midrule
     \multirow{2}*{GoodNews}        & Image     & 445,259   & 19,448        & 24,461 \\
    ~                               & Article   & 254,906   & 18,741        & 23,402 \\
    \bottomrule
    \end{tabular}
\end{table}

\section{Experiments}

\begin{table*}
\caption{Entity-aware captioning performance on the BreakingNews and GoodNews datasets. `NewsUsage' refers to at which level the model uses the corresponding news article.}
\vspace{-10pt}
    \label{tab:Caption_Experiments}
    \footnotesize
    \centering
    \begin{tabular}{p{0.10\linewidth}p{0.15\linewidth}p{0.1\linewidth}|p{0.04\linewidth}p{0.04\linewidth}p{0.05\linewidth}p{0.04\linewidth}|p{0.04\linewidth}p{0.04\linewidth}p{0.05\linewidth}p{0.04\linewidth}}
    \toprule
    \multirow{2}*{\textbf{Type}} & \multirow{2}*{\textbf{Models}}  & \multirow{2}*{\textbf{NewsUsage}} &  \multicolumn{4}{c|}{\textbf{BreakingNews}}  &\multicolumn{4}{c}{\textbf{GoodNews}}   \\
    ~ & ~ & ~ & Bleu-4 & Meteor & Rouge-L & CIDEr & Bleu-4 & Meteor & Rouge-L & CIDEr \\
    \midrule
    \multirow{6}*{Two-stage} & SAT\cite{DBLP:conf/icml/XuBKCCSZB15}+CtxIns & - & 0.63 &4.02 & 10.63 & 13.94 &0.38 &3.56 & 10.50 & 12.09 \\
    ~ & Att2in2\cite{rennie2017self}+CtxIns  & - &0.69 & 3.98 & 10.23 &14.09  &0.74 & 4.17 & 11.58 &13.71 \\
    ~ & TopDown\cite{DBLP:conf/cvpr/00010BT0GZ18}+CtxIns & - &0.38 & 3.83 & 10.46 &14.89 &0.70 & 3.81 & 11.09 &13.38 \\
    ~ & Ramisa \etal \cite{ramisa2018breakingnews}+CtxIns & Article Level&0.71 & 4.33 & 10.41 &14.62  &0.89 & 4.45 & 12.09 &15.35 \\
    ~ & Biten \etal \cite{DBLP:conf/cvpr/BitenGRK19}+CtxIns &Sentence Level &0.67 &4.06 & 10.66 & 14.63 &0.69 & 4.14 & 11.70 & 13.77 \\
    ~ & Biten \etal \cite{DBLP:conf/cvpr/BitenGRK19}+SAttxIns & Sentence Level &0.90 & 4.41 & 11.42 & 17.28 &0.76 & 4.02 & 11.44 & 13.69  \\
    \midrule
    \multirow{2}*{End-to-end} & LSTM\_BPE \cite{Tran2020Tell} & Article Level  &1.30  &5.19 & 11.13 &9.91 &1.92 & 5.44 &13.48 &13.43 \\
    ~ & ICECAP & Word Level  &\textbf{1.71} &\textbf{6.00} & \textbf{14.33} &\textbf{25.74} &\textbf{1.96} &\textbf{6.01} &\textbf{15.70} &\textbf{26.08} \\
    \bottomrule
    \end{tabular}
\end{table*}

\begin{table}
    \caption{Named entity Generation performance (F1 score) on the BreakingNews and GoodNews datasets.}
    \vspace{-10pt}
    \label{tab:NES_Experiments}
    \footnotesize
    \centering
    \begin{tabular}{p{0.15\linewidth}p{0.3\linewidth}|p{0.2\linewidth}p{0.2\linewidth}}
    \toprule
    \texttt{Type} & \textbf{Models}  & \textbf{BreakingNews} & \textbf{GoodNews}  \\
    \midrule
    \multirow{6}*{Two-stage} & SAT\cite{DBLP:conf/icml/XuBKCCSZB15}+ CtxIns &7.61  &6.09\\
    ~ & Att2in2\cite{rennie2017self} + CtxIns  &8.07  &6.53 \\
    ~ & TopDown\cite{DBLP:conf/cvpr/00010BT0GZ18} + CtxIns  &12.26 &6.87\\
    ~ & Ramisa \etal \cite{ramisa2018breakingnews} + CtxIns &8.45  &7.54 \\
    ~ & Biten \etal\cite{DBLP:conf/cvpr/BitenGRK19}   +  CtxIns &8.77   &7.19 \\
    ~ & Biten \etal \cite{DBLP:conf/cvpr/BitenGRK19}  + SAttxIns &12.60 &7.97\\
    \midrule
    \multirow{2}*{End-to-end} & LSTM\_BPE\cite{Tran2020Tell}  &6.36  & 8.42 \\
    ~ & ICECAP  &\textbf{17.33}  &\textbf{12.03} \\
    \bottomrule
    \end{tabular}
\end{table}

\subsection{Experimental Setup}
\noindent\textbf{Datasets.}
We conduct experiments on two news image caption datasets, the BreakingNews \cite{ramisa2018breakingnews} and GoodNews  \cite{DBLP:conf/cvpr/BitenGRK19}.
In BreakingNews dataset, the average article length is 25 sentences.
Each image contains one groundtruth image caption written by expert journalists. There are 15 words and 2 named entities on average for each caption.
In GoodNews dataset, the average article length is 22 sentences and there are only one annotated image caption for each image by journalists.
Each caption contains 14 words and 3 named entities on average.
Partition statistics of the two datasets are shown in Table \ref{tab:dataset statistic}.

\noindent\textbf{Implementation Details.} 
We use SpaCy toolkit \cite{honnibal2017spacy} to recognize named entities in the sentence.
The max length of caption is set to 31 for both two datasets. 
The vocabulary size for BreakingNews and GoodNews is 15098 and 32768, respectively.
We use the $5^{th}$ layer of ResNet152\cite{DBLP:conf/cvpr/HeZRS16} as our region-level visual representations $V$.
We use the top 10 sentences ranked by our retrieval model as the relevant sentences for the downstream captioning model.
During training, the scalar $\gamma$ in Eq~(\ref{final_loss}) is set to 0.2. 

\noindent\textbf{Evaluation Metric.}
We use BLEU \cite{DBLP:conf/acl/PapineniRWZ02}, METEOR \cite{DBLP:conf/wmt/DenkowskiL14}, ROUGE-L \cite{lin2004rouge} and CIDEr \cite{DBLP:conf/cvpr/VedantamZP15} as the evaluation metrics for entity-aware image captioning. CIDEr is most suitable for this task because it puts more importance on rarer words such as named entities by tf-idf weights.  
To further evaluate models' ability of generating named entities for news images, we compute F1 score between named entities set in groundtruth caption and those in generated caption. Exact lexicon match is applied for comparing two named entities.
However, to be noted, the F1 score is only an optimistic metric to reflect named entity generation because the contexts of named entities are not taken into account for the correctness of a generated named entity.

\begin{table*}[!htbp]
    \footnotesize
    \centering
    \caption{Entity-aware captioning performance of some models with different sets of sentences. `Info Concentration' means whether to use our information concentration module to select relevant sentences. ICECAP-M means that ICECAP drops the word-level matching and uses word-level attention weights to insert named entities (WAttIns).}
    \vspace{-8pt}
    \begin{tabular}{p{0.12\linewidth}p{0.10\linewidth}p{0.12\linewidth}|p{0.04\linewidth}p{0.03\linewidth}p{0.05\linewidth}p{0.04\linewidth}|p{0.04\linewidth}p{0.04\linewidth}p{0.05\linewidth}p{0.04\linewidth}}
    \toprule
    \multicolumn{3}{c|}{\textbf{Models}}  & \multicolumn{4}{c|}{\textbf{BreakingNews}}  &\multicolumn{4}{c}{\textbf{GoodNews}}   \\
    Template Captioning & Entity Insertion  & Info Concentration & Bleu-4 & Meteor & Rouge-L & CIDEr & Bleu-4 & Meteor & Rouge-L & CIDEr \\
    \midrule
    \multirow{3}*{Biten \etal \cite{DBLP:conf/cvpr/BitenGRK19}}     & CtxIns  & $\times$ &0.67 &4.06 & 10.66 & 14.63  &0.69 & 4.14 & 11.70 & 13.77\\
    ~       & CtxIns & $\checkmark$  &  0.74 & 4.16 & 10.72 & 14.27 &0.55 & 3.75 & 11.21 & 12.40 \\
    ~  & SAttxIsn & $\checkmark$  &0.78 & 4.02 & 10.64 & 13.62 &0.54 & 3.52 & 11.04 & 11.72  \\
    \midrule
    ICECAP-M  & WAttIns & $\checkmark$  &1.54 &5.73 &13.74 &22.59 &1.65 & 5.36 &14.05 &19.79  \\
    \bottomrule
    \end{tabular}
    \label{tab:Caption_relevant_sent_Experiments}
\end{table*}

\subsection{Comparison with State-of-the-Arts}
We compare ICECAP model with some state-of-the-art approaches. 

\vspace{4pt}
\noindent\textbf{Two-stage Models:}
These approaches follow a two-stage pipeline that firstly generates template captions and then inserts named entities.
To generate the template caption, the first type of models are image-only models, which are the state-of-the-art general image captioning models, including SAT \cite{DBLP:conf/icml/XuBKCCSZB15}, Att2in2 \cite{rennie2017self} and TopDown \cite{DBLP:conf/cvpr/00010BT0GZ18}. These models only use the image information to generate template caption.
The second type of models are news image captioning models, which incorporate article information to generate template captions. Ramisa \etal \cite{ramisa2018breakingnews} leverage global article embedding in the caption decoding. Biten \etal \cite{DBLP:conf/cvpr/BitenGRK19} dynamically attend to sentence embeddings in the article.
To insert named entities, a general method is the Context-based Insertion (CtxIns) \cite{DBLP:conf/cvpr/BitenGRK19}. Ramisa \etal \cite{ramisa2018breakingnews} further propose a Sentence-level Attention Insertion (SAttIns) method.

\vspace{4pt}
\noindent\textbf{End-to-end Models:}
Alasdair \etal \cite{Tran2020Tell} use byte-pair encoding(BPE) \cite{DBLP:conf/acl/SennrichHB16a} which can overcome the OOV problem. 
For fair comparison, we choose the basic model LSTM\_BPE in their work as our baseline. It encodes the article to a single vector by averaging the glove word embeddings and applies 4 LSTM layers to decode the byte-pair tokens with region-level image attention. 

Table \ref{tab:Caption_Experiments} presents the entity-aware news image captioning performance on two datasets.
First, our ICECAP model outperforms state-of-the-art models on all metrics, especially CIDEr score on both two datasets. It demonstrates the effectiveness of fine-grained information concentrated article encoding and named entity matching mechanism. Second, We can see that the article information plays an important role for news image captioning by comparing models without using any article encoding to models with article encoding. Besides, for two-stage models that use article information, performance of Ramisa \etal \cite{ramisa2018breakingnews}+CtxIns and Biten \etal  \cite{DBLP:conf/cvpr/BitenGRK19}+CtxIns are not consistent on different datasets, which might suggests that encoding articles at either article level or sentence level alone is not sufficient to capture article information.

We further evaluate the named entity generation performance in Table \ref{tab:NES_Experiments}.
We find that ICECAP performs much better than both two-stage models and end-to-end model with byte-pair encoding. It suggests that with the help of word-level matching mechanism, ICECAP has better ability to select named entities from the corresponding articles.

\subsection{Ablation Studies on Information Concentration Module}

\paragraph{\textbf{Are the selected sentences sufficient to concentrate relevant information in the whole article?}}

We demonstrate the effectiveness of our information concentration module via comparing the downstream captioning performance of using the selected sentences and the whole article.
 
Since it is computational inefficient for our ICECAP model to process the whole article, we choose the sentence-level model Biten \etal \cite{DBLP:conf/cvpr/BitenGRK19} in the comparison. 
As shown in Table \ref{tab:Caption_relevant_sent_Experiments}, Biten \etal \cite{DBLP:conf/cvpr/BitenGRK19}+CtxIns with selected sentences achieves comparable performance with the one using the whole article. It shows that there is not much valuable information loss from our information concentration module.

We also show that after sentence selection, our proposed captioning model can make better use of the selected sentence.
We use a variant of ICECAP - ICECAP-M for fair comparison with Biten \etal \cite{DBLP:conf/cvpr/BitenGRK19}, which drops the word-level matching and uses word-level attention weights to insert named entities (WAttIns). 
Results are presented in Table \ref{tab:Caption_relevant_sent_Experiments}. It further proves that encoding sentences at fine-grained word level is necessary to make good use of the selected sentences.

\begin{figure}[!htbp]
    \centering
    \includegraphics[width=0.9\linewidth]{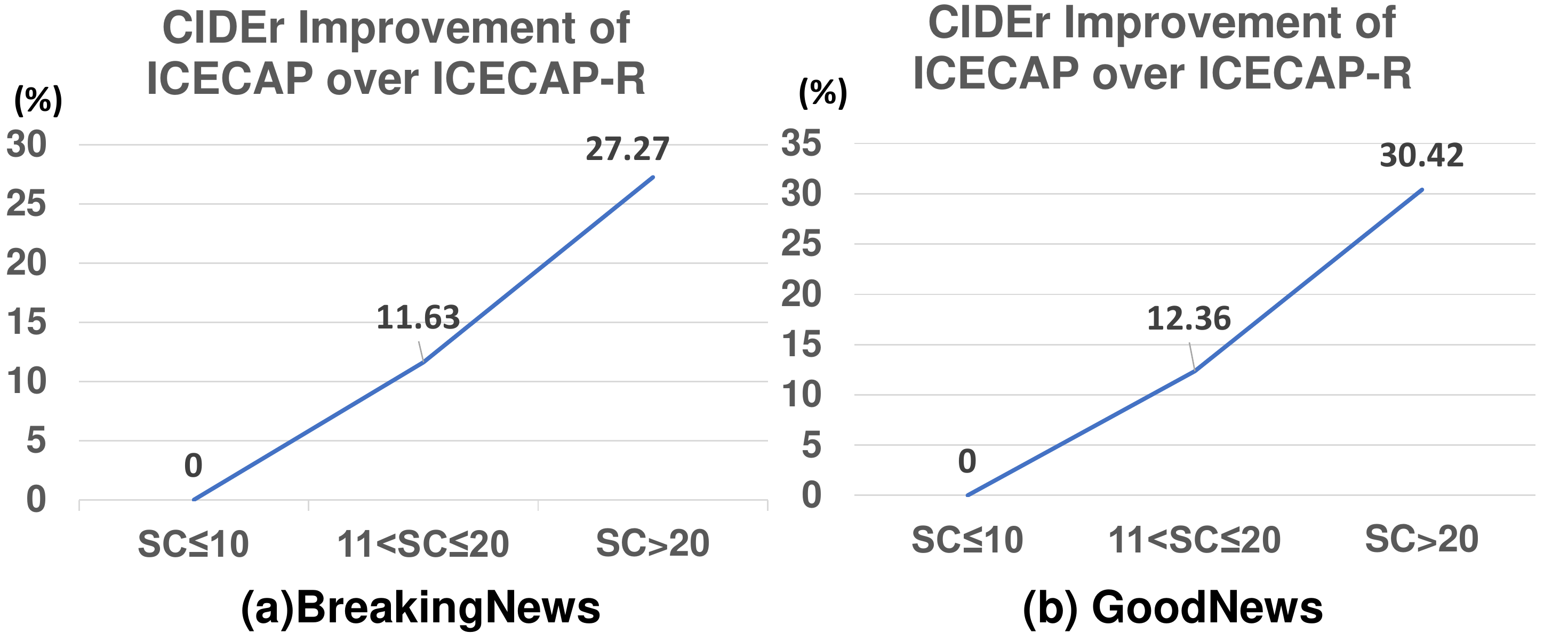}
    \vspace{-10pt}
    \caption{Improvement on CIDEr given by information concentration module in different subset of test data. `SC' refers to the sentence count of the corresponding article.}
    \label{fig:retrieval_impro}
\end{figure}

To know how much the information concentration module contributes to the entity-aware captioning performance of ICECAP, we compare ICECAP with ICECAP-R. 
There is only one difference between ICECAP and ICECAP-R that during inference, ICECAP-R randomly chooses 10 sentences from the corresponding article for the downstream entity-aware image captioning.
Because of the uncertainty of the random selection, we run ICECAP-R three times with different randomly chosen sentences and use their average performance as the final performance for ICECAP-R. 
Figure \ref{fig:retrieval_impro} shows the improvement of ICECAP over ICECAP-R on CIDEr in different subsets of test data, which are split according to the sentence count(SC) of the corresponding article.  
We find the improvement given in `SC$>$20' subset is more than twice the improvement in `11$<$SC$\leq$20' subset on both datasets. This indicates that our information concentration module brings more gain when the article has more sentences and therefore more noise for entity-aware caption generation. Besides, the `SC$>$20' subset accounts for 49.45\% and 35.19\% of BreakingNews and GoodNews test set, respectively, which further supports the necessity of our information concentration module.

\begin{figure}[!htbp]
    \centering
    \includegraphics[width=0.9\linewidth]{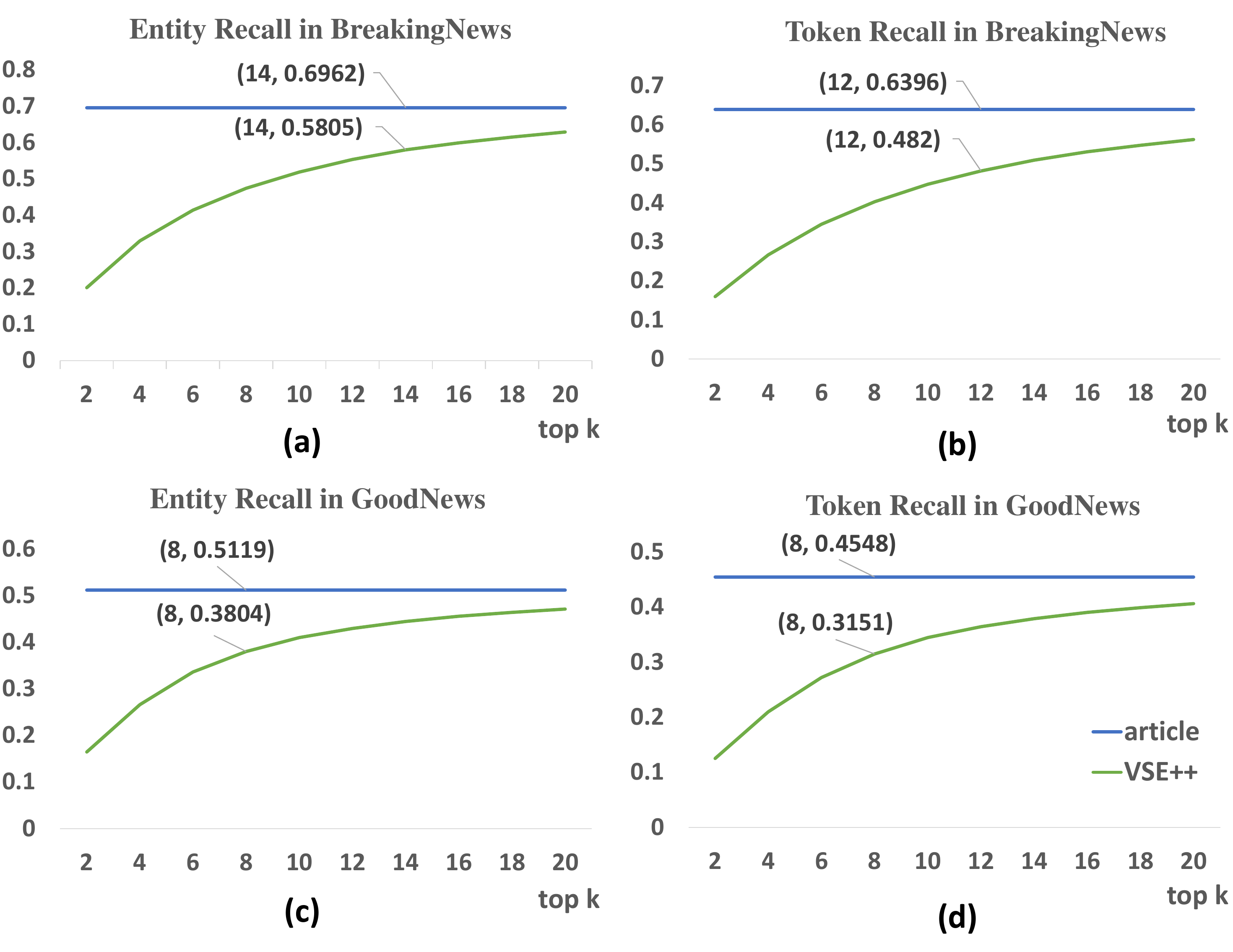}
    \vspace{-10pt}
    \caption{Entity recall and informative token recall in BreakingNews and GoodNews datasets. `article' refers to all sentences in the article. `VSE++' refers to the top k relevant sentences ranked by our pre-trained  retrieval model.}
    \label{fig:recall}
\end{figure}

\begin{figure}[!htbp]
    \centering
    \includegraphics[width=0.92\linewidth]{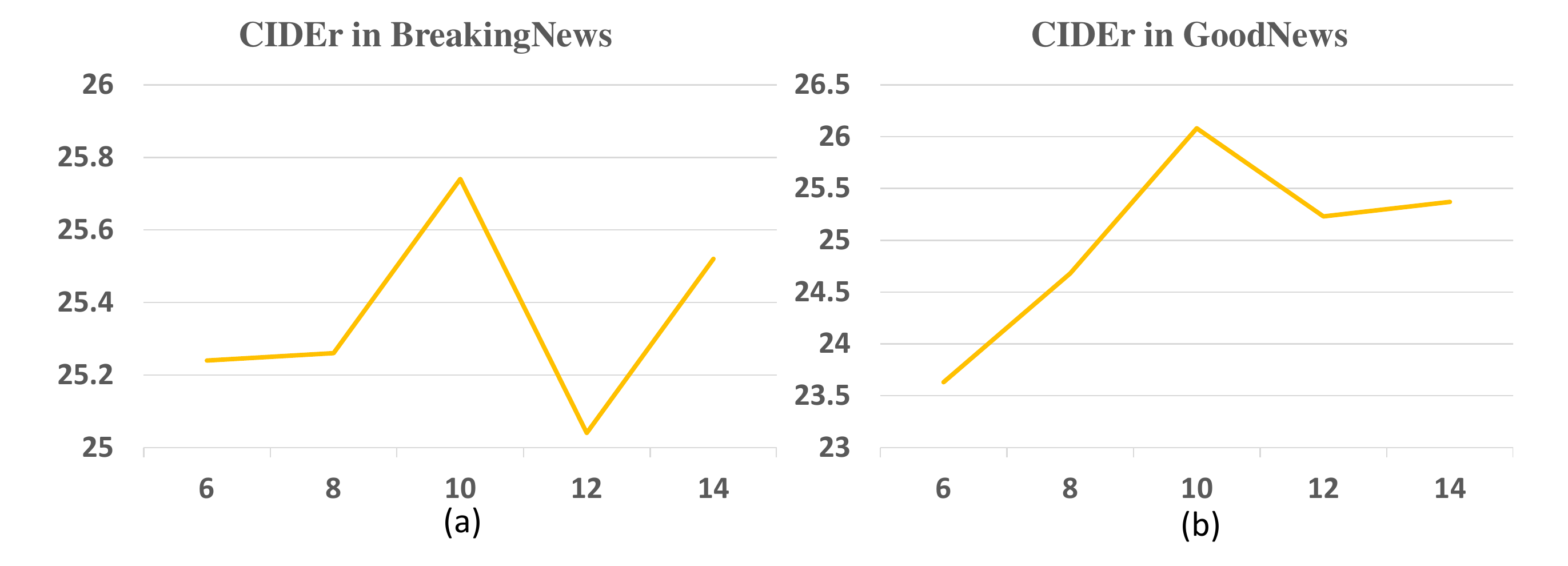}
    \vspace{-10pt}
    \caption{CIDEr scores of ICECAP with different R values in test set of BreakingNews and GoodNews datasets.}
    \label{fig:finetuneR}
\end{figure}

\paragraph{\textbf{How to set the number of relevant sentences to be selected?}}
Our information concentration module selects top R most relevant sentences for our downstream entity-aware captioning model. These sentences are expected to contain as much key information as the whole article does, while greatly reduce the number of words. To measure how much key information is contained in the sentences, we use the the recall for informative tokens (tokens excluding stop words) and named entities in captions as the metrics. 
Considering that the average numbers for informative tokens and named entities in a news caption are about 11 and 3, we think the gap of $0.15$ on recall compared with the whole article is acceptable. As shown in Figure \ref{fig:recall}, the first k to enter the 0.15 gap for informative tokens and named entities on BreakingNews (GoodNews) dataset are 14 (8) and 12 (8), respectively. By averaging these four numbers, we set the R to 10, which is less than half of the average sentence count of an article in both datasets.

Besides the pre-experiment analysis, we also carry out experiments of using different R ranging from 6 to 14 to see the influence of R on the performance of ICECAP. As shown in Figure \ref{fig:finetuneR}, ICECAP with R set to 10 achieves the best performance on both two datasets. It shows selecting less sentences may lose valuable information. Selecting more sentences may generate a 
word sequence that is too long to be well encoded by LSTM.

\begin{figure*}
    \centering
    \includegraphics[width=0.9\linewidth]{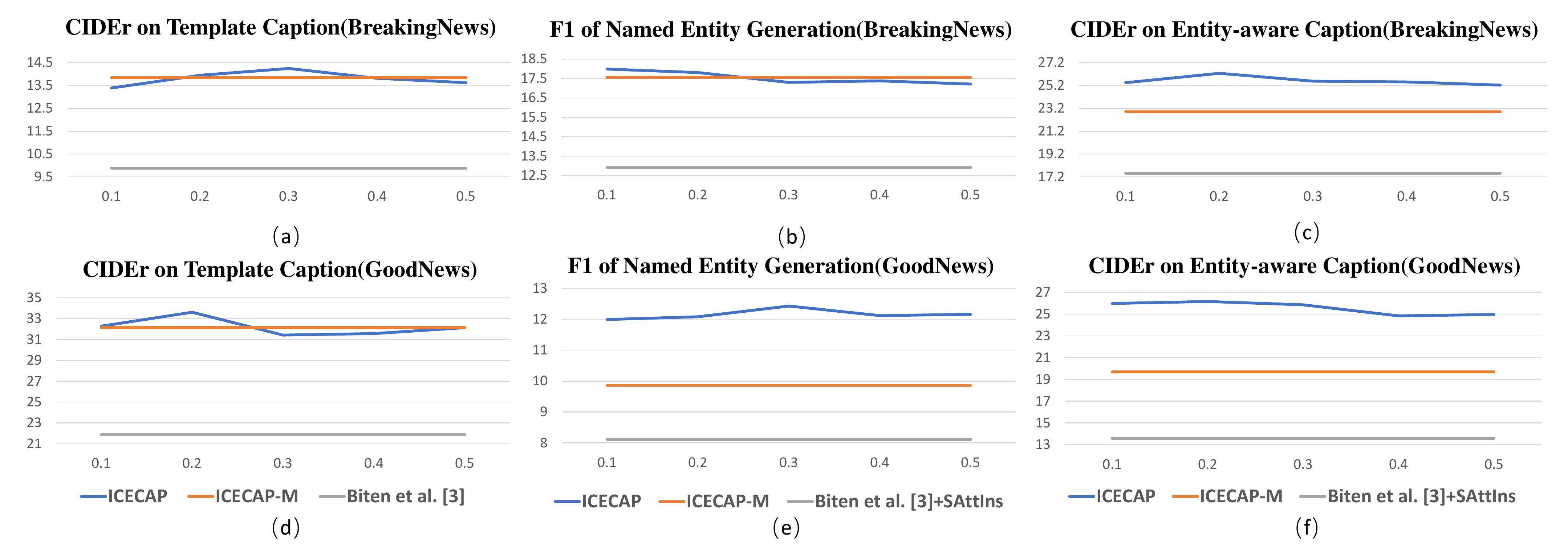}
    \vspace{-10pt}
    \caption{Trade-off between event description and named entity generation. The x-axis of each chart is the trade-off scalar $\gamma$.}
    \label{fig:tradeoff}
\end{figure*}

\subsection{Ablation Studies on Entity-aware Captioning Module}
We further conduct experiments to verify the effectiveness of different components in the entity-aware caption generation module.

\paragraph{\textbf{Entity-distinguished Embedding.}}
First, as shown in Table \ref{tab:ablation study all}, ICECAP outperforms ICECAP-S on both datasets. It shows that our entity-distinguished embedding indeed helps our model to distinguish different entities with the same entity category and generate better entity-aware image captions.

\paragraph{\textbf{Global Article Encoding.}}
Second, ICECAP performs slightly better than ICECAP-G on CIDEr with the help of global article feature, which indicates that adding the overall article feature at the initial decoding step is beneficial as it may help compensate textual information loss from the information concentration step.

\paragraph{\textbf{Named Entity Matching.}}
Third, ICECAP outperforms ICECAP-M on both two datasets. Specially, on GoodNews dataset, there is a significant decline for the entity-aware captioning and the entity generation. It indicates that our word-level matching layer can contribute more to finding appropriate named entities in the case of more training data.

\begin{table}[!htbp]
    \footnotesize
    \centering
    \caption{Ablation study on Entity-aware Captioning Module. ICECAP-S drops serial number embedding when encoding local word-level features. ICECAP-G drops the global article feature. ICECAP-M drops the word-level matching layer and uses word-level attentions to select named entities.}
    \vspace{-10pt}
    \begin{tabular}{p{0.16\linewidth}|p{0.15\linewidth}|p{0.08\linewidth}p{0.05\linewidth}p{0.10\linewidth}p{0.06\linewidth}|p{0.1\linewidth}}
    \toprule
    \multirow{2}*{\textbf{Dataset}} & \multirow{2}*{\textbf{Models}}  & \multicolumn{4}{c|}{\textbf{Entity-aware Captioning}}  & \multirow{2}*{Entity F1}\\
    ~ & ~   & Bleu-4 & Meteor & Rouge-L & CIDEr & ~ \\
    \midrule
    \multirow{4}*{BreakingNews} & ICECAP-S  &1.71 & 5.89 & 14.02 & 24.27 & 16.01 \\
    ~ & ICECAP-G  &1.45 & 5.51 & 13.73 & 24.16 & 16.29\\
    ~ & ICECAP-M  &1.54 &5.73 &13.74 &22.59 & 17.08\\
    ~ & ICECAP  &\textbf{1.71} &\textbf{6.00} & \textbf{14.33} &\textbf{25.74} & \textbf{17.33} \\
    \midrule
    \multirow{4}*{GoodNews} & ICECAP-S  &1.71 &5.51 &14.62 &22.28 & 10.62 \\
    ~ & ICECAP-G  &1.60 &5.27 &14.14 &21.40 & 10.94 \\
    ~ & ICECAP-M  &1.65 & 5.36 &14.05 &19.79 & 9.93\\
    ~ & ICECAP  &\textbf{1.96} &\textbf{6.01} &\textbf{15.70} &\textbf{26.08} & \textbf{12.03} \\
    \bottomrule
    \end{tabular}
    \label{tab:ablation study all}
    \vspace{-8pt}
\end{table}

\begin{figure*}[!htbp]
    \centering
    \includegraphics[width=0.9\linewidth]{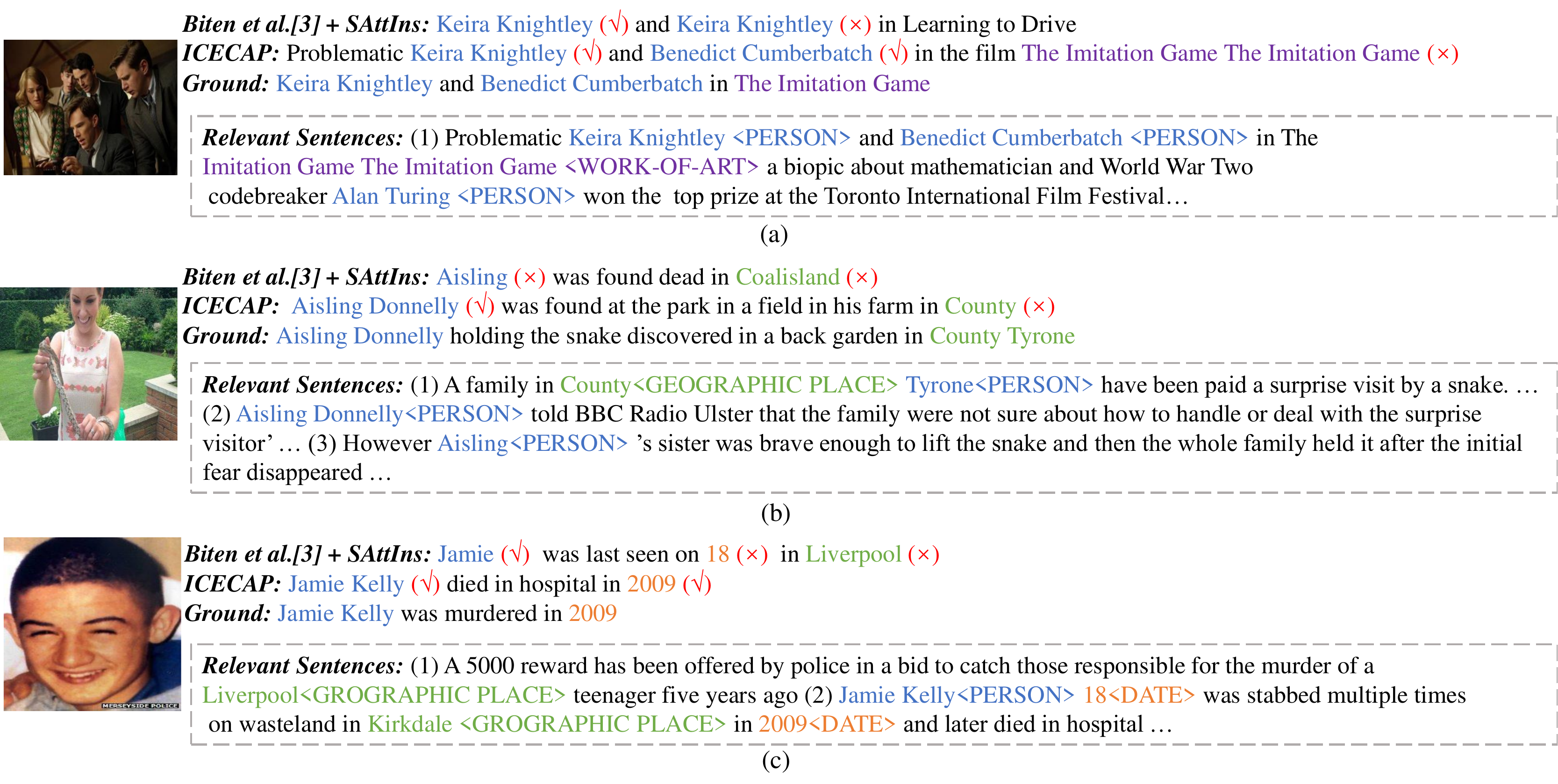}
    \vspace{-8pt}
    \caption{Qualitative Results of ICECAP and Biten \etal \cite{DBLP:conf/cvpr/BitenGRK19}+SAttIns}
    \label{fig:case}
\end{figure*}

\subsection{Trade-off between Event Description and Named Entity Generation}
\label{sec:trade-off}
A good caption for news image should not only contain appropriate named entities but also describe the relevant news event accurately. To evaluate the performance on the event description, the influence of named entity should be excluded. Thus the CIDEr on the template caption is an appropriate evaluation metric. In ICECAP, $\rm{L_p}$ and $\rm{L_e}$ are designed for event description and named entity selection, respectively. We use a scalar $\gamma$ to trade off these two aspects. To determine an appropriate $\gamma$, we train ICECAP models with different $\gamma$ values and compare their performance on the validation set. Besides, we compare these models with ICECAP-M and Biten \etal \cite{DBLP:conf/cvpr/BitenGRK19}+SAttIns on the performance of template captioning, named entity generation and entity-aware captioning. 

First, as shown in Figure \ref{fig:tradeoff}(a) and Figure \ref{fig:tradeoff}(d), ICECAP-M outperforms Biten \etal \cite{DBLP:conf/cvpr/BitenGRK19} significantly on generating the template caption. This indicates that information concentration module and our word-level attention mechanism contribute a lot for the event description.
Besides, ICECAP with a relative small value of $\gamma$ can achieve comparable or better performance on template captioning compared with ICECAP-M. 
This might be because the supervision on named entity selection is not accurate enough.

Second, as shown in Figure \ref{fig:tradeoff}(b) and Figure \ref{fig:tradeoff}(e), ICECAP models with appropriate $\gamma$ values outperform ICECAP-M and Biten \etal \cite{DBLP:conf/cvpr/BitenGRK19}+SAttIns on named entity generation. This proves that optimizing the entity selection loss $L_e$ indeed helps the model select appropriate named entities, especially on large dataset.

Last, Figure \ref{fig:tradeoff}(c) and Figure \ref{fig:tradeoff}(f) shows that ICECAP models with different $\gamma$ values (from 0.1 to 0.5) all generate better entity-aware captions than ICECAP-M and Biten \etal \cite{DBLP:conf/cvpr/BitenGRK19}+SAttIns. This indicates that ICECAP can well balance the event description and the named entity generation. We finally set $\gamma$ to 0.2, with which ICECAP achieves the best performance on entity aware captioning on both datasets.

\vspace{-6pt}
\subsection{Case Analysis}
We present some examples from the test sets in Figure \ref{fig:case}. As shown in Figure \ref{fig:case}(a), Biten \etal \cite{DBLP:conf/cvpr/BitenGRK19}+SAttIns accurately find the relevant sentences for the two person placeholders by sentence-level attention weights.
However, it can not justify which person name in the sentence is appropriate for each person placeholder without word-level clues. Thus it reuses an identical entity name `Keira Knightley' twice in the final caption. With the help of the word-level matching mechanism, ICECAP accurately locates the appropriate named entities in the relevant sentences for both person placeholders. 
In the example shown in Figure \ref{fig:case}(b), the third sentence which is most relevant with the event doesn't contain the target person name `Aisling Donnelly'. The only person name `Aisling' in this sentence refers to another person, the brother of `Aisling Donnelly'. Biten \etal \cite{DBLP:conf/cvpr/BitenGRK19}+SAttIns locates this sentence through the sentence-level attention weights and incorrectly inserts the person name `Aisling' in its final caption. Due to our word-level matching mechanism, ICECAP is able to look through every word in the relevant sentences and successfully select the person name from a less relevant sentence. 
Besides more accurate named entity generation, ICECAP also performs better on the event description.
As we can see from the figure \ref{fig:case}(c), the image shows a boy who died of murder. With the word-level attention mechanism, ICECAP is capable of looking through every word in the relevant sentences and locates one of the key phrases `die in the hospital'. 
Thus, it generates a caption more close to the background of the image, compared with the one generated by Biten \etal \cite{DBLP:conf/cvpr/BitenGRK19}+SAttIns.
Furthermore, the mistakes made by ICECAP in the first two examples reveal that our method still suffers from the performance of named entity recognition during data pre-processing, which is a common shortage in template-based methods.

\section{Related Work}
\vspace{4pt}
\noindent\textbf{Image Captioning.}
In recent years, there have been many models based on neural networks \cite{rennie2017self,DBLP:conf/icml/XuBKCCSZB15,DBLP:conf/cvpr/00010BT0GZ18,DBLP:conf/cvpr/LuXPS17,DBLP:conf/cvpr/VinyalsTBE15} proposed for general image captioning. 
These methods focus on describing the major objects and their surface relations but fail to express the related background or concrete named entities about the image.

\vspace{4pt}
\noindent\textbf{Entity-aware Image Captioning.}There have been some research endeavours to generate informative and entity-aware captions, for example, utilizing external sources from the web \cite{DBLP:conf/emnlp/LuWHJC18,DBLP:conf/acl/ZhaoSLS19}, or finding supplementary text information from associated news articles \cite{feng2012automatic,DBLP:journals/tip/TariqF17,ramisa2018breakingnews,DBLP:conf/cvpr/BitenGRK19, Tran2020Tell}. 
To leveraging the article information, some works \cite{feng2012automatic,DBLP:journals/tip/TariqF17} propose to use the probability model to learn the relation between the image and the article.
Ramisa \etal \cite{ramisa2018breakingnews} represent each article with a single vector and uses a LSTM layer to fuse the image feature and the article feature to decode the caption. To overcome the out-of-vocabulary problem caused by named entities, Alasdair \etal \cite{Tran2020Tell} propose to represent entity-aware caption with byte-pair encoding. Biten \etal \cite{DBLP:conf/cvpr/BitenGRK19} propose to generate a template caption first and then insert named entities during post-processing. Biten \etal \cite{DBLP:conf/cvpr/BitenGRK19} is the work most related to us. 
There are three main differences between their method and ours: First, we apply a information concentration to concentrate on the sentences relevant to the image. Second, our caption generation model can attend to different words in the relevant sentences to get more fine-grained textual information. Third, our caption generation model is end-to-end. It jointly generates the template caption and selects the most appropriate named entities with a match mechanism.

\vspace{4pt}
\noindent\textbf{Image-Text Cross-modal Retrieval.}
Image-Text Cross-modal  Retrieval focus on retrieving the most relevant images(s) given an caption or retrieving the most relevant caption(s) given an image. 
Some attention mechanisms\cite{DBLP:conf/cvpr/HuangWW17,DBLP:conf/cvpr/NamHK17} are designed to compute the similarity between an image and a caption.
When learning the visual-semantic embeddings, VSE++\cite{DBLP:conf/bmvc/FaghriFKF18} focuses on the hardest negatives to calculate the triplet ranking loss rather than all negatives.

\section{Conclusion}
To generate entity-aware captions for news images, we propose a novel framework ICECAP, which progressively concentrates on the associated article from coarse level to fine-grained level. 
The proposed model first generates coarse concentration on relevant sentences through cross-modality retrieval. Then the model further focuses on fine-grained words and generates the entity-aware image caption in the end-to-end manner. It looks through every word in the relevant sentences by a word-level attention layer at each decoding step and uses a word-level match mechanism to choose appropriate named entities from relevant sentences. 
Our ICECAP model achieves state-of-the-art results on both BreakingNews and GoodNews datasets.

\begin{acks}
This work was supported by National Natural Science Foundation of China (
No.61772535) and Beijing Natural Science Foundation (No.4192028).
\end{acks}

\bibliographystyle{ACM-Reference-Format}
\bibliography{cap}

\end{document}